\documentclass{article}

\usepackage{arxiv}

\usepackage[utf8]{inputenc} 
\usepackage[T1]{fontenc}    
\usepackage{hyperref}       
\usepackage{url}            
\usepackage{booktabs}       
\usepackage{amsfonts}       
\usepackage{amsmath}
\usepackage{nicefrac}       
\usepackage{microtype}      

\usepackage{hyperref}
\usepackage{url}
\usepackage{subcaption}
\usepackage{multirow} 
\usepackage{graphicx}
\usepackage{natbib}
\usepackage{doi}
\usepackage{cleveref}       
\usepackage{lipsum}         

\usepackage{makecell}
\usepackage{tabularx}

\title{No Trust Issues Here: A Technical Report on the Winning Solutions for the Rayan AI Contest}

\date{}

\newif\ifuniqueAffiliation
\uniqueAffiliationtrue

\author{ Ali Nafisi \\
	Department of Computer Engineering \\
	Bu-Ali Sina University\\
	\texttt{a.nafisi@eng.basu.ac.ir} \\
	\And
	Sina Asghari\thanks{Equal contribution.} \\
	Department of Computer Science\\
	Iran University of Science and Technology \\
	\texttt{sina\_asghari@mathdep.iust.ac.ir} \\
	\And
	Mohammad Saeed Arvenaghi\footnotemark[1] \\
	Department of Computer Science\\
	Iran University of Science and Technology \\
	\texttt{m\_arvenaghi@mathdep.iust.ac.ir} \\
    \And
	Hossein Shakibania \\
	Department of Computer Science\\
	Technical University of Darmstadt \\
	\texttt{hossein.shakibania@stud.tu-darmstadt.de} \\
}

\renewcommand{\headeright}{Technical Report}
\renewcommand{\undertitle}{}
\renewcommand{\shorttitle}{No Trust Issues Here}

\begin{document}
\maketitle

\begin{abstract}

    This report presents solutions to three machine learning challenges developed as part of the Rayan AI Contest: compositional image retrieval, zero-shot anomaly detection, and backdoored model detection. In compositional image retrieval, we developed a system that processes visual and textual inputs to retrieve relevant images, achieving 95.38\% accuracy and ranking first with a clear margin over the second team. For zero-shot anomaly detection, we designed a model that identifies and localizes anomalies in images without prior exposure to abnormal examples, securing second place with a 73.14\% score. In the backdoored model detection task, we proposed a method to detect hidden backdoor triggers in neural networks, reaching an accuracy of 78\%, which placed our approach in second place. These results demonstrate the effectiveness of our methods in addressing key challenges related to retrieval, anomaly detection, and model security, with implications for real-world applications in industries such as healthcare, manufacturing, and cybersecurity. Code for all solutions is available online.\footnote{\url{https://github.com/safinal/rayan-ai-contest-solutions}}
    
\end{abstract}

\keywords{Compositional Retrieval \and Zero-Shot Anomaly Detection \and Backdoored Model Detection}

\section{Introduction}

Traditional machine learning systems often struggle with scenarios where data distributions shift or when input data is incomplete or noisy. This report addresses three such challenges, each focusing on improving the reliability, flexibility, and security of machine learning systems. As the demand for intelligent systems grows across industries, the need for models that can process both visual and textual data, detect anomalies in unseen distributions, and safeguard against adversarial manipulations becomes increasingly critical.

The solutions detailed in this report were developed for the Rayan AI Contest\footnote{\url{https://ai.rayan.global}}, an international competition hosted by Sharif University of Technology focused on Trustworthy AI. The contest attracted significant global participation, with over 4,600 contestants from more than 60 countries forming 730 teams during the screening phase. From this pool, 100 teams were invited to the main phase, and ultimately, more than 40 teams successfully participated with at least one valid submission. To address the competition's objectives, we focused on three distinct tasks: compositional image retrieval, zero-shot anomaly detection, and backdoored model detection. 

The first task involves developing a system capable of retrieving images based on multi-modal queries. The second focuses on detecting anomalies in industrial and medical settings, and the third aims to identify security vulnerabilities, specifically backdoor attacks, in deep learning models. 

The evaluation of these solutions is based on a set of standard metrics, such as accuracy and F1 score, to assess both the quality and robustness of the approaches. Our team, \textit{No Trust Issues Here}, secured the 1st place overall by achieving state-of-the-art performance across all three tracks. Specifically, we ranked 1st in compositional image retrieval with a score of 95.38\%, 2nd in zero-shot anomaly detection with 73.14\%, and 2nd in backdoored model detection with 78\%. The final leaderboard ranking was determined by Total Points, calculated as:

$$\text{Total Points} = \sum_{i=1}^{3} (101 - \text{rank}_i)$$

where $\text{rank}_i$ denotes the team's rank in the $i$-th challenge. Our cumulative 298 points placed us at the top of the final leaderboard, as shown in Table \ref{tab:leaderboard-full}.

\begin{table}[h!]
    \centering
    \small
    \caption{Final leaderboard showing the top 10 teams.}
    \label{tab:leaderboard-full}
    \begin{tabular}{c l c c c c}
        \toprule
        \multirow{2}{*}{\textbf{Rank}} & \multirow{2}{*}{\textbf{Team}} & \multicolumn{3}{c}{\textbf{Rank}} & \multirow{2}{*}{\textbf{\thead{Total\\Points}}} \\
        \cmidrule(lr){3-5}
        & & \thead{Compositional\\Image Retrieval} & \thead{Zero-Shot\\Anomaly Detection} & \thead{Backdoored\\Model Detection} & \\
        \midrule
        1 & \textbf{No Trust Issues Here (Our Team)} & \textbf{1} & \textbf{2} & \textbf{2} & \textbf{298} \\
        2 & Pileh & 4 & 1 & 6 & 292 \\
        3 & AI Guardians of Trust & 2 & 6 & 4 & 291 \\
        4 & AIUoK & 9 & 3 & 5 & 286 \\
        5 & red\_serotonin & 3 & 7 & 9 & 284 \\
        6 & Persistence & 7 & 11 & 3 & 282 \\
        7 & GGWP & 6 & 9 & 11 & 277 \\
        8 & Tempest & 11 & 5 & 13 & 274 \\
        9 & AlphaQ & 5 & 10 & 16 & 272 \\
        10 & Cogniverse & 20 & 13 & 12 & 258 \\
        \bottomrule
    \end{tabular}
\end{table}

\section{Compositional Image Retrieval}
\label{sec:q5}

\begin{figure}
    \centering
    \includegraphics[width=0.7\linewidth]{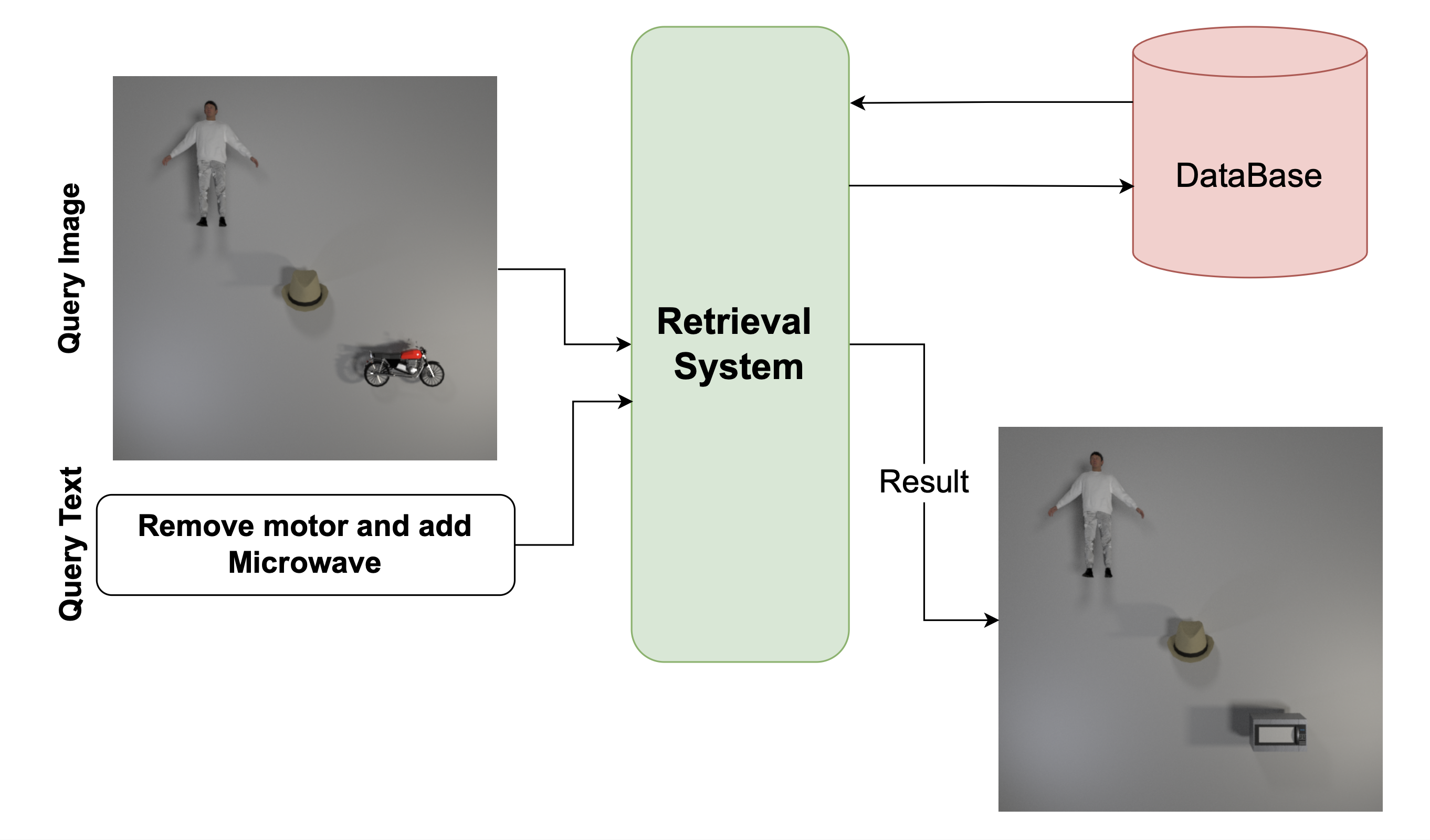}
    \caption{An overview of the Compositional Image Retrieval problem.}
    \label{fig:fig1}
\end{figure}

The objective of the Compositional Image Retrieval challenge is to develop an intelligent search system capable of processing multi-modal queries. The system must accept a reference image containing a visual scene and a textual modification description (e.g., "Remove motor and add Microwave") as input. The goal is to identify and retrieve the most relevant image from a database that reflects the visual changes described in the text. Figure \ref{fig:fig1} visualizes the compositional retrieval problem.

This task presents a unique challenge in that it requires the model to understand both the visual content of the reference image and the semantic intent of the textual instructions to perform accurate retrieval, all while adhering to strict resource constraints such as a 4GB model size limit and a prohibition on large language models or object detection models during inference.

\subsection{Methodology}

To address the challenge of compositional retrieval without relying on prohibited heavy architectures, we propose a two-stage pipeline: \textit{Token Classification} and \textit{Compositional Embedding Arithmetic}. This approach provides a lightweight method for modifying visual representations using textual instructions while remaining within the competition constraints.
All code associated with our implementation is available at the provided repository.
\footnote{\url{https://github.com/safinal/compositional-image-retrieval}}

\subsubsection{Token Classification for Semantic Parsing}

\begin{figure}
    \centering
    \includegraphics[width=\linewidth]{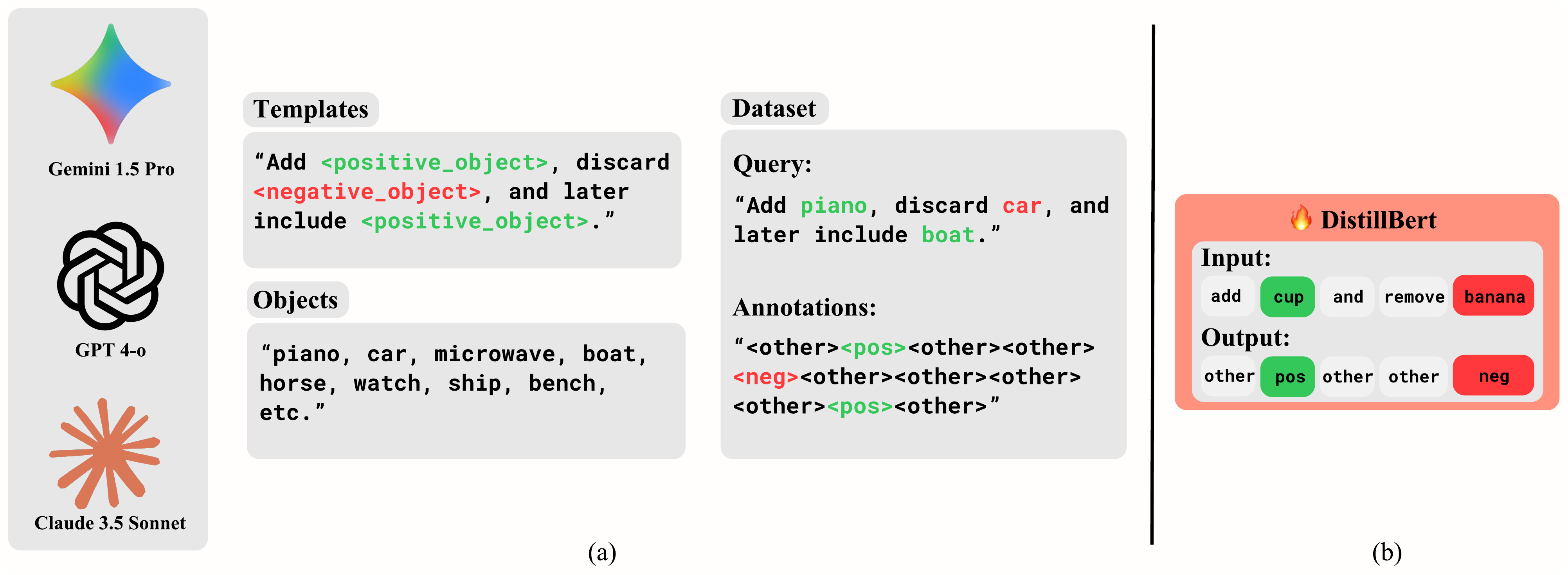}
    \caption{Overview of the token classification module. (a) Dataset curation using templated queries instantiated with LLM-generated object vocabulary. (b) Fine-tuned DistilBERT model for token-level classification into \textit{positive}, \textit{negative}, and \textit{other} categories, using the curated dataset.}
    \label{fig:q5_graphical_abstract}
\end{figure}

The first stage of our pipeline is the Token Classification model, which is responsible for interpreting the query text and identifying specific modifications (i.e., which objects to add or remove), as illustrated in Figure~\ref{fig:q5_graphical_abstract}. This task is framed as a token classification problem where the model classifies tokens into three categories:
\begin{itemize}
    \item Positive ($pos$): Objects to be added to the scene.
    \item Negative ($neg$): Objects to be removed from the scene.
    \item Other: Irrelevant tokens (stopwords, punctuation, verbs).
\end{itemize}

To create training data, we generated a synthetic dataset using publicly available LLM interfaces (Gemini 1.5 Pro, GPT-4o, and Claude 3.5 Sonnet). No API access was used. We built 636 template sentences containing placeholders for positive and negative objects. We additionally generated a vocabulary of 1,018 unique object names. Each template was instantiated 15 times by replacing the placeholders with random objects from the list, producing 9,540 labeled queries.

Labels were assigned automatically. For a query such as ``add apple and remove banana'', the sequence is annotated as <other> <positive\_object> <other> <other> <negative\_object>. This procedure ensures consistent detection of objects to add and remove.

We fine-tuned \texttt{distilbert-base-uncased} \citep{sanh2019distilbert}. DistilBERT was selected due to its small size and adequate performance for this restricted classification task. The model was trained for 20 epochs with a weight decay of $0.01$, 500 warm-up steps, and standard token-level cross-entropy. This stage outputs two sets: detected positive objects and detected negative objects.

\subsubsection{Compositional Embedding Arithmetic for Retrieval}

The second stage retrieves the target image by modifying the vector representation of the query image according to the objects identified in the query text. We evaluated multiple models from OpenCLIP \citep{ilharco_gabriel_2021_5143773} and selected ViTamin-L-384 \citep{chen2024vitamin} due to its superior performance in preliminary experiments. To remain within resource limits, we fine-tune only the visual head (\texttt{model.feature\_extractor.visual.head}) while keeping the rest of the backbone fixed.

We optimize the retrieval model using the InfoNCE loss \citep{oord2018representation}. The objective is to align the composed query vector with the vector of the ground-truth target image. Let $\mathbf{v}_{\text{query}} \in \mathbb{R}^d$ be the query image embedding and $\mathbf{t}_{\text{pos}}, \mathbf{t}_{\text{neg}} \in \mathbb{R}^d$ be the embeddings of the objects to be added and removed, respectively. In practice, the modification vectors are computed by encoding the text prompts ``a photo of a [object]'' using the model's text encoder. During training, this encourages the model to learn a consistent direction in the embedding space for each object addition or removal.

Training utilizes the AdamW optimizer (learning rate $10^{-4}$, weight decay $0.01$) over five epochs with a loss temperature of $0.07$. We employ a CosineAnnealingWarmRestarts scheduler with $T_0 = 5$ and $T_{\text{mult}} = 2$. Because the dataset contains repeated occurrences of the same target image, we implement a custom sampler to ensure that each batch contains unique target images, which is required for the InfoNCE formulation.

During inference, the Token Classifier identifies the sets of positive objects $\mathcal{P}$ and negative objects $\mathcal{N}$ from the query text. We encode these to obtain the corresponding embedding vectors $\mathbf{t}_{\text{pos}}^{(i)}$ and $\mathbf{t}_{\text{neg}}^{(j)}$. Given the query image vector $\mathbf{v}_{\text{query}}$, the predicted target vector $\mathbf{v}_{\text{target}}$ is computed as:

$$\mathbf{v}_{\text{target}} = \mathbf{v}_{\text{query}} + \sum_{i \in \mathcal{P}} \mathbf{t}_{\text{pos}}^{(i)} - \sum_{j \in \mathcal{N}} \mathbf{t}_{\text{neg}}^{(j)}$$

The system then computes the cosine similarity between $\mathbf{v}_{\text{target}}$ and all image vectors in the database, returning the image with the highest similarity score.

\begin{figure}
    \centering
    \includegraphics[width=1.0\linewidth]{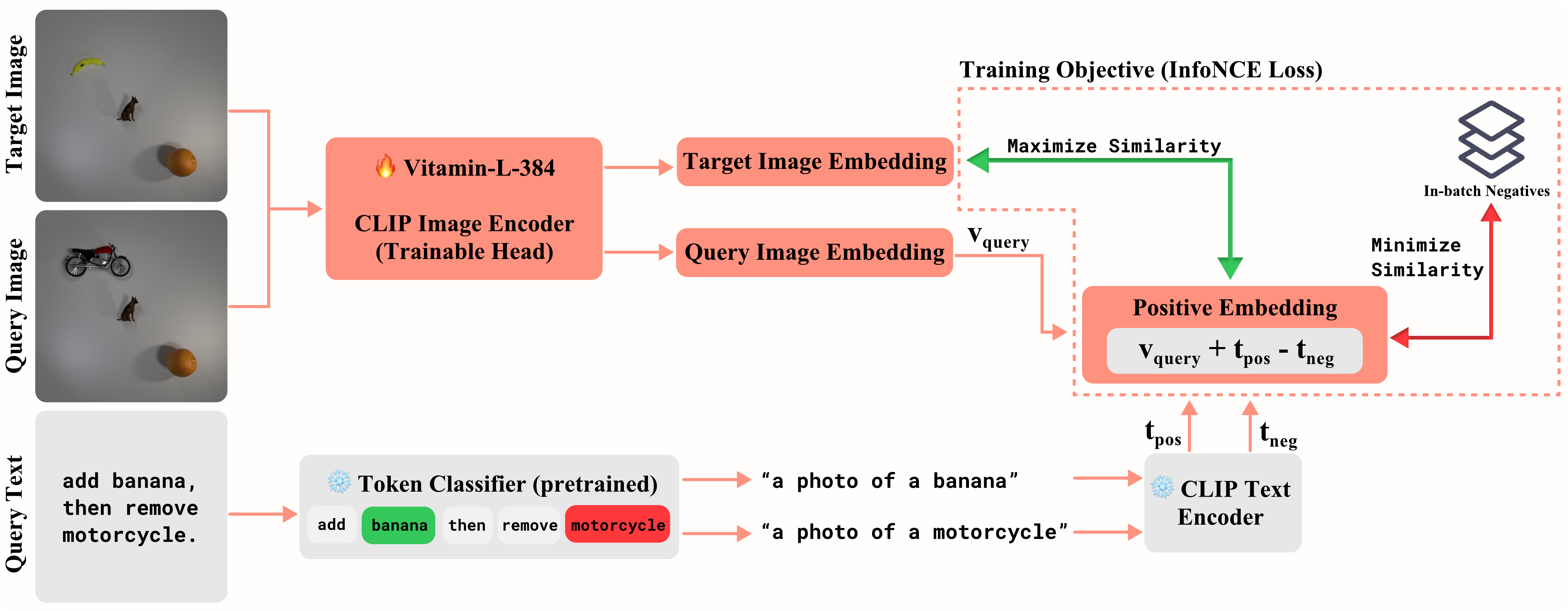}
    \caption{Overview of the proposed compositional image retrieval method. The system parses textual modifications using our fine-tuned token classifier and applies embedding arithmetic to adjust the reference image representation before retrieval.}
    \label{fig:q5_graphical_abstract_2}
\end{figure}

This method does not rely on object detection, captioning models, or large language models during inference. By operating directly in the latent space, the approach remains efficient while enabling precise compositional retrieval, 
as illustrated in Figure~\ref{fig:q5_graphical_abstract_2}.

\subsection{Experimental Results}
The evaluation was conducted on a test set designed to assess generalization capability through diverse text descriptions and novel visual combinations. The primary metric for performance was \textit{Top-1 Accuracy}.

Our solution demonstrated superior performance, achieving the highest accuracy among all participating teams. The model successfully handled complex queries involving simultaneous addition and removal of objects. Table \ref{tab:table1} shows the top 10 leaderboard rankings for this problem.

\begin{table}[h]
    \centering
    \caption{Top 10 Leaderboard Rankings for the Compositional Image Retrieval Challenge.}
    \label{tab:table1}
    \begin{tabular}{clc}
        \toprule
        \textbf{Rank} & \textbf{Team} & \textbf{Accuracy (\%)} \\
        \midrule
        \textbf{1} & \textbf{No Trust Issues Here (Our Team)} & \textbf{95.38} \\
        2 & AI Guardians of Trust & 88.59 \\
        3 & red\_serotonin & 85.45 \\
        4 & Pileh & 84.61 \\
        5 & AlphaQ & 82.89 \\
        6 & GGWP & 80.97 \\
        7 & Persistence & 80.63 \\
        8 & Synapse & 78.66 \\
        9 & AIUoK & 78.32 \\
        10 & fatem17 & 75.76 \\
        \bottomrule
    \end{tabular}
\end{table}

With a final accuracy of 95.38\%, our approach outperformed the second-place entry by a significant margin of 6.79\%, validating the effectiveness of combining explicit semantic parsing with vector arithmetic in latent space.

\section{Zero-Shot Anomaly Detection}
\label{sec:q6}
Reliability in machine learning models is often hindered by the closed-world assumption, where test data are presumed to match training distributions. In real-world scenarios, this assumption frequently fails as models encounter outlier samples. To address this, we focused on the task of "Anomaly Detection," specifically within the challenging setting of zero-shot learning.

Unlike conventional settings that rely on normal samples for training, the zero-shot approach requires the model to operate without seeing any data—normal or anomalous—from the test-time distribution during its training phase. The only source of knowledge available regarding the test distribution is the implicit information contained within the unlabeled test set itself.

\begin{figure}[h]
    \centering


    \includegraphics[width=0.70\linewidth]{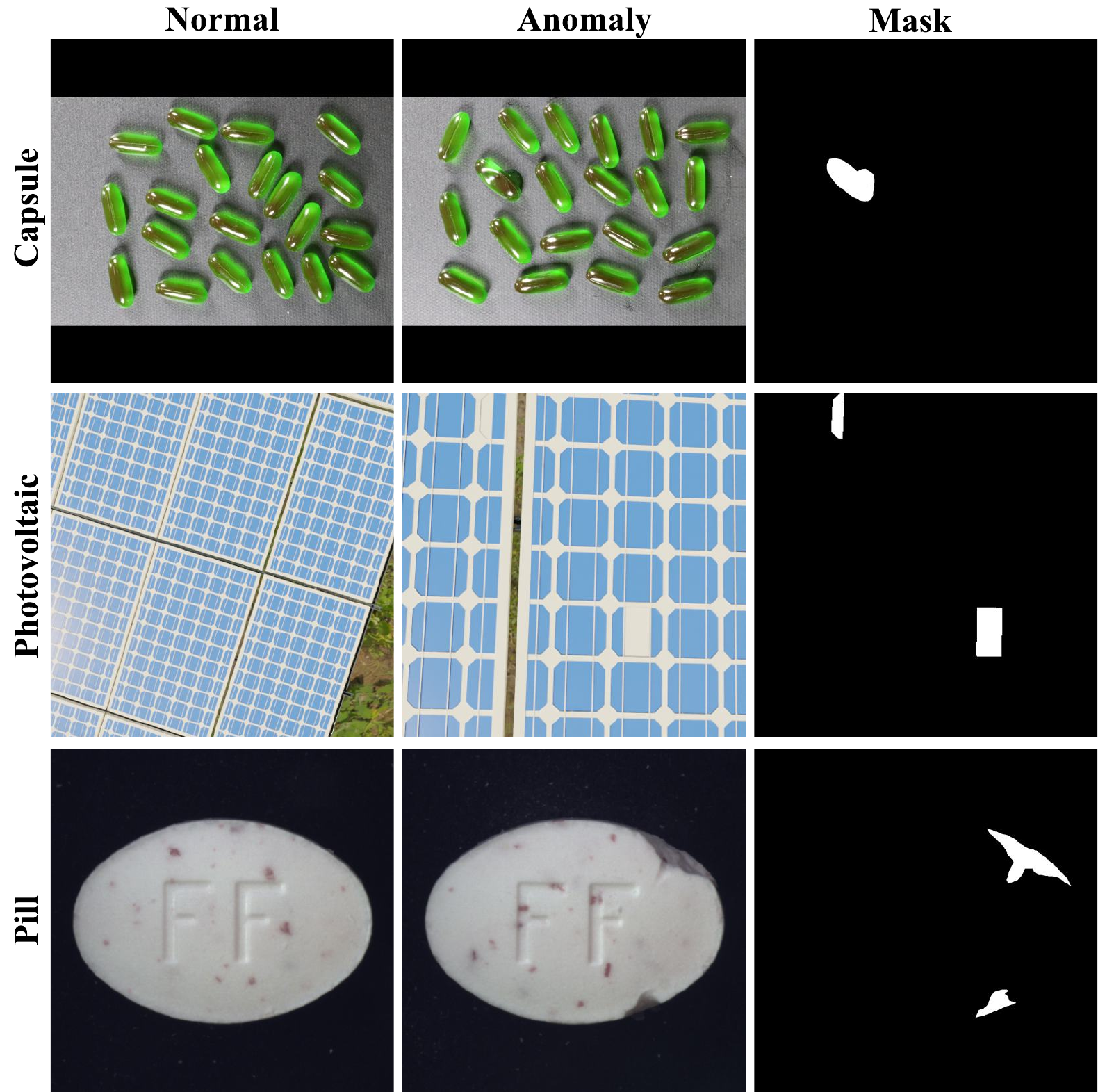}
    \caption{Examples of normal images, anomalous images, and anomaly masks across three datasets: capsules, photovoltaic modules, and pills.}
    \label{fig:q6_examples}
\end{figure}

The problem is formulated as follows: given a set of test images $D_{test}=\{I_{test}^1, ..., I_{test}^{n_{test}}\}$, and potentially a set of auxiliary training images $D_{aux}$ distinct from the test distribution, the goal is to determine whether a test image is anomalous and localize the specific anomalous regions. The output for each sample consists of two components:
\begin{itemize}
    \item Classification Score ($s_{img} \in \mathbb{R}$): An image-level scalar indicating the degree of anomaly.
    \item Segmentation Score ($s_{pix} \in \mathbb{R}^{H \times W}$): A pixel-level map localizing the defect.
\end{itemize}
This task targets both industrial and medical domains, requiring the model to generalize across eight industrial classes and two medical classes. In these domains, the distribution shift between normal and abnormal data is primarily a covariate shift, as they share the same semantics (e.g., a normal pill vs. a broken pill).
Figure~\ref{fig:q6_examples} shows examples of normal and anomalous images from three datasets (capsules, photovoltaic modules, and pills). For each dataset, we display a normal sample, an anomalous sample, and the corresponding anomaly mask.

\subsection{Methodology}

After reviewing several methods, including AnomalyCLIP \citep{AnomalyCLIP}, we selected the MuSc method \citep{li2024musc} because it achieved the strongest results among available zero-shot anomaly detection approaches. MuSc is also the first zero-shot method that does not require class descriptions, unlike CLIP-based models.

Our approach follows the MuSc framework \citep{li2024musc}, which performs zero-shot anomaly detection using only unlabeled test images. MuSc is based on the observation that normal image patches tend to recur across many test samples, whereas anomalous patches are uncommon. Patch features are extracted with a Vision Transformer, and each patch is scored by comparing it to patches from the remaining images.

MuSc contains three main components. The first is the Local Neighborhood Aggregation with Multiple Degrees (LNAMD), which aggregates patch features at different neighborhood sizes to capture anomalies at multiple scales. The second is the Mutual Scoring Mechanism (MSM), which assigns patch-level anomaly scores by measuring how often a patch finds similar counterparts in the test set, using an interval-average over the lowest similarity values to reduce noise. The third component, Re-scoring with Constrained Image-level Neighborhood (RsCIN), refines image-level anomaly scores by constructing a constrained neighborhood graph and enforcing consistency on images with similar global features.

Building on this framework, we performed a systematic search over hyperparameters and selected settings that consistently improved performance on both image-level and pixel-level metrics. Our final configuration used feature layers $\{5, 11, 17, 23\}$, $r \in \{1, 3\}$, $k_{score} \in \{1, 8, 9\}$. We also adopted a combined model design, using \citep{oquab2023dinov2} (dinov2-vitl14) for segmentation and \citep{radford2021learning} (ViT-L-14-336) for classification.

During evaluation, we identified one difficult class (photovoltaic modules) with weak performance. We conducted targeted error analysis on these cases and applied class-specific adjustments that improved their results without harming performance on other classes. We performed error analysis using only the three industrial classes provided for local development and then tuned hyperparameters on that validation set. No labels or statistics from the official blind evaluation classes were used for tuning.

We introduced a classical computer vision filter to produce a binary mask used for post-processing the anomaly maps. This step was motivated by a limitation we observed in MuSc. Because the method relies only on patch dissimilarities and has no prior information or class description, it implicitly assumes that normal samples have consistent textures. 
In some classes, such as photovoltaic modules in Figure~\ref{fig:q6_examples}, normal images contain background regions (grass, soil) that differ strongly from the object surface. These surrounding areas can be incorrectly treated as anomalies.
To reduce this effect, the binary mask defines a narrow margin around the main object. Pixels with mask value 1 keep their original anomaly values, and all other pixels are set to the minimum of the anomaly map. This post-processing step improved stability and enhanced pixel-level localization across multiple classes.

For image-level classification, instead of taking only the highest patch anomaly score per image, we obtained better results by combining the two highest patch scores. The second-highest value is included with a weight of 0.25, which improved robustness compared to a single maximum score.
All code associated with our implementation is available at the provided repository.\footnote{\url{https://github.com/safinal/zeroshot-anomaly-detection}}

\subsection{Experimental Results}

The performance of our submission was evaluated using a rigorous weighted average of seven distinct metrics, assessing precision, recall, and specificity at both the image and pixel levels. The inference phase was executed on an NVIDIA GeForce RTX 4090 GPU with a maximum time limit of 3 hours.

The final evaluation score was computed using a weighted average of all metrics:

$$Final~score=\frac{\sum_{metric \in Metrics}\mathcal{W}_{metric}\times metric}{\sum_{metric \in Metrics}\mathcal{W}_{metric}}$$

The specific weights assigned to each metric were:
\begin{itemize}
    \item Image-level: $w_{img\_AUROC}=1.2$, $w_{img\_AP}=1.1$, $w_{img\_F1}=1.1$
    \item Pixel-level: $w_{pix\_AUROC}=1.0$, $w_{pix\_AUPRO}=1.4$, $w_{pix\_AP}=1.3$, $w_{pix\_F1}=1.3$
\end{itemize}

Table \ref{tab:table2} shows the leaderboard for this problem. Our team, \textit{No Trust Issues Here}, achieved the second highest performance in the challenge. By leveraging the logic detailed in the methodology section, our solution attained a final score of 73.14\%, securing the 2nd rank on the leaderboard. This performance demonstrated superior generalization across the diverse industrial and medical classes provided in the blind test set.

\begin{table}[h!]
    \centering
    \caption{Top 10 Leaderboard Rankings for the Zero-Shot Anomaly Detection Challenge.}
    \begin{tabular}{clc}
        \toprule
        \textbf{Rank} & \textbf{Team} & \textbf{Score (\%)} \\
        \midrule
        1 & Pileh & 74.92 \\
        2 & \textbf{No Trust Issues Here (Our Team)} & \textbf{73.14} \\
        3 & AIUoK & 72.98 \\
        4 & Tempest & 70.88 \\
        5 & AI Guardians of Trust & 66.29 \\
        6 & red\_serotonin & 63.51 \\
        7 & CortexAI & 62.35 \\
        8 & GGWP & 62.25 \\
        9 & AlphaQ & 62.15 \\
        10 & Persistence & 61.62 \\
        \bottomrule
    \end{tabular}
    \label{tab:table2}
\end{table}

\section{Backdoored Model Detection}
\label{sec:q7}

The objective of this challenge is to develop a discrimination function capable of distinguishing between clean and backdoored neural networks. As deep neural networks are increasingly deployed in mission-critical applications, the risk of attackers poisoning training data to embed imperceptible backdoors has become a significant security concern.

The threat model involved attackers capable of data poisoning and training influence to create stealthy, label-consistent, or sample-specific triggers. The defense mechanism was required to operate without prior knowledge of the attack type or trigger structure, relying only on the model parameters and a small set of clean samples (1\% of the train dataset).

\subsection{Methodology}
Our approach builds on the Mm-Bd method \citep{wang2024mm}, which reaches approximately 60\% accuracy, and extends it by introducing feature-space optimization, improved initialization, image resizing, and a refined statistical detection rule. These modifications address the constraint of model-only detection and lead to substantially higher accuracy.
All code associated with our implementation is available at the provided repository.
\footnote{\url{https://github.com/safinal/backdoored-model-detection}}

\subsubsection{Activation Optimization}
Since the original triggers are unavailable, we attempt to reverse-engineer the model's sensitivity to potential trigger patterns. The pipeline operates on the PreActResNet18 architecture \citep{he2016identity} as follows:
\begin{enumerate}
    \item \textbf{Initialization:} For each class $c$, we sample a batch of clean images from the verification set. Before optimization, we resize all input images to $128 \times 128$, as this resolution consistently improved accuracy compared to the default preprocessing.

    \item \textbf{Forward Pass \& Activation Extraction:} The resized images are passed through the model up to \texttt{model.layer1[0]} to produce the initial feature tensor. We found that initializing from these real features gives better results than the random tensor initialization used in Mm-Bd.

    \item \textbf{Optimization:} We treat the extracted activations as trainable variables and freeze all other activations and layers. We then optimize the activations to maximize the model's confidence for a target class. This searches for a latent trigger in the feature space; after evaluating all layers of PreActResNet18, we identified \texttt{model.layer1[0]} as the optimal point for exposing target-class over-sensitivity in backdoored models.
    
    \item \textbf{Maximal Confidence Scoring:} After the optimization, we compute a maximal confidence score for each class, penalized by activation magnitudes of other classes to ensure specificity.
\end{enumerate}

We use hyperparameters that were stable across datasets: 100 optimization steps, batch size 30, an initial learning rate of $10^{-1}$ decayed to $10^{-3}$, and 10 warmup steps.

\subsubsection{Statistical Outlier Detection}
Our detection hypothesis is that clean models show balanced confidence across classes, while backdoored models exhibit a single target class with unusually high maximal confidence.

Rather than using a fixed threshold, we apply a statistical anomaly test:
\begin{itemize}
    \item \textbf{Distribution Fitting:} We collect the maximal confidence scores across classes, exclude the maximum, and fit the remaining scores to an exponential distribution. We tested all SciPy \citep{2020SciPy-NMeth} candidate distributions and found the exponential distribution produced the best separation between clean and backdoored models.
    
    \item \textbf{Hypothesis Testing:} We treat the maximum score as a candidate outlier and compute its $p$-value under the fitted distribution.

    \item \textbf{Decision Rule:} If the $p$-value is below $0.08$, the model is classified as backdoored; otherwise, it is classified as clean.
\end{itemize}

This statistical approach provides robustness across datasets and attack types while maintaining the constraint of using only limited, fixed parameters.

\begin{table}[h!]
    \centering
    \caption{Top 10 Leaderboard Rankings for the Backdoored Model Detection Challenge.}
    \begin{tabular}{clc}
        \toprule
        \textbf{Rank} & \textbf{Team} & \textbf{Accuracy (\%)} \\
        \midrule
        1 & AUTs & 79 \\
        2 & \textbf{No Trust Issues Here (Our Team)} & \textbf{78} \\
        3 & Persistence & 74 \\
        4 & AI Guardians of Trust & 72 \\
        5 & AIUoK & 70 \\
        6 & Pileh & 67 \\
        7 & My Team & 66 \\
        8 & Unknown & 66 \\
        9 & red\_serotonin & 65 \\
        10 & DevNull & 65 \\
        \bottomrule
    \end{tabular}
    \label{tab:accuracy-leaderboard}
\end{table}

\subsection{Experimental Results}
The evaluation was conducted on a private test dataset where models were trained on various image classification datasets (e.g., CIFAR10, MNIST) using different backdoor attack types. The evaluation environment utilized a single Nvidia RTX 4090 GPU with a strict 1-minute computation limit per data sample.
The primary metric for success was Accuracy, defined as the ratio of correct predictions (identifying backdoored vs. clean models) to the total number of data samples.

Table \ref{tab:accuracy-leaderboard} summarizes the results. Our proposed method demonstrated high efficacy, achieving the 2nd rank overall. We achieved an accuracy of 78\%, finishing only a single percentage point behind the winning team. These results show that the method detects several backdoor types without prior attack information.

\section{Conclusion}
This report addressed three critical challenges in machine learning: compositional image retrieval, zero-shot anomaly detection, and backdoored model detection. In the compositional retrieval task, we developed a system that successfully combined visual and textual inputs to retrieve relevant images, achieving first place with a top accuracy of 95.38\%. For zero-shot anomaly detection, we implemented a method that detected and localized anomalies in unseen data, securing second place with an accuracy of 73.14\%. In the backdoored model detection task, our approach identified backdoor attacks in deep neural networks, achieving a 78\% accuracy, placing second in the competition. These results demonstrate the effectiveness of our methodologies in addressing each problem's specific challenges.

The solutions developed in this report have significant implications for real-world applications, including medical imaging, industrial quality control, and security. By enabling models to process multi-modal queries, detect unseen anomalies, and identify backdoored networks, our work contributes to building more reliable, robust, and secure AI systems. Future work could focus on further improving model generalization, handling more complex real-world scenarios, and refining methods to detect novel attack strategies. Ultimately, these challenges highlight the need for continuous innovation to ensure that machine learning systems can operate effectively and securely across diverse environments.

\clearpage
\pagebreak

\bibliographystyle{unsrtnat}
\bibliography{references}  






\end{document}